# Chapter 5

# Explainable AI in Spatial Analysis

**Ziqi Li**[1,2] (**Ziqi.Li@fsu.edu**)

[1]Department of Geography, Florida State University, Tallahassee, FL 32306 USA
[2]The Spatial Data Science Center, Florida State University, Tallahassee, FL 32306 USA

## Abstract

This chapter discusses the opportunities of eXplainable Artificial Intelligence (XAI) within the realm of spatial analysis. A key objective in spatial analysis is to model spatial relationships and infer spatial processes to generate knowledge from spatial data, which has been largely based on spatial statistical methods. More recently, machine learning offers scalable and flexible approaches that complement traditional methods and has been increasingly applied in spatial data science. Despite its advantages, machine learning is often criticized for being a black box, which limits our understanding of model behavior and output. Recognizing this limitation, XAI has emerged as a pivotal field in AI that provides methods to explain the output of machine learning models to enhance transparency and understanding. These methods are crucial for model diagnosis, bias detection, and ensuring the reliability of results obtained from machine learning models. This chapter introduces key concepts and methods in XAI with a focus on Shapley value-based approaches, which is arguably the most popular XAI method, and their integration with spatial analysis. An empirical example of county-level voting behaviors in the 2020 Presidential election is presented to demonstrate the use of Shapley values and spatial analysis with a comparison to multi-scale geographically weighted regression. The chapter concludes with a discussion on the challenges and limitations of current XAI techniques and proposes new directions.

## Introduction

Spatial modeling has long been a cornerstone of spatial analysis, which provides essential tools to understand spatial relationships underlying data and to predict future spatial scenarios. Traditionally, spatial modeling has largely focused on statistical approaches. Widely applied methods include spatial econometric models (e.g., spatial lag model, spatial error model, spatial Durbin model), spatially varying coefficients models (such as geographically weighted regression and eigenvector filtering), Kriging-based models, and Bayesian spatial models, among others (Anselin, 1988; Fotheringham et al.,

2002; Griffth, 2002; Hengl et al., 2007; Lindgren and Rue, 2015; Lee, 2013). These statistical models are advantageous in which they can explicitly specify spatial effects such as spatial autocorrelation and spatial heterogeneity. They also offer inferential frameworks that helps quantify uncertainties and infer underlying processes. However, most statistical approaches often come with strong data and model requirements, such as assumptions of data distribution, model linearity, and scale, stationarity and isotropy of spatial effects, which may simplify spatial relationships in the real world. Moreover, these models can be challenging to specify and are computationally intensive, particularly with moderate-to-large datasets, limiting their scalability and utilities where increasingly fine-scale, large-coverage spatial data are available. These limitations put challenges in discovering hidden combinations of spatial, non-linear, and interaction effects that are more likely to occur in the real world but may go undetected due to intrinsic model specification, estimation, or computational issues. In fact, many social and natural behaviors and processes are highly nonlinear and are influenced by spatial context which include social interactions, economic activities, urbanization processes, climate systems, and ecological dynamics, among others (Helbing, 2010; Rial et al., 2004; Lek et al., 1996).

Over the past decade, machine learning has emerged as a powerful technique in spatial data science (Li, 2019; Janowicz et al., 2020; Hu et al., 2023; Gao et al., 2024). Machine learning presents significant opportunities for spatial data modeling by addressing many of the limitations inherent in traditional spatial and geostatistical approaches. These models excel at capturing non-linear interactions, require fewer assumptions, and are highly efficient in processing high-dimensional and high-volume data. As universal approximators, machine learning models do not rely on specific model specifications and data assumptions; instead, they learn functions and effects directly from the data (Sonoda and Murata, 2017). Additionally, they are scalable to large datasets and can seamlessly incorporate diverse data types (e.g., numerical and categorical) and distributions. The integration of machine learning in the geospatial domain has given rise to the emerging field of Geospatial AI (GeoAI) that applies artificial intelligence to address spatial problems (Gao et al., 2023). GeoAI has shown significant potential and promise across a wide range of applications, including weather prediction (Lam et al., 2023), spatial interpolation (Zhu et al., 2020), spatial object classification and detection (Li et al., 2020), and many social and human geography studies (Wang et al., 2024).

However, machine learning models are often considered black boxes because their internal workings are not easily understandable. This is primarily due to their complexity, which often involves non-linear transformations, high-dimensional feature interactions, model ensembling, and iterative training processes. These mechanisms result in a vast number of parameters, making them difficult to interpret and posing challenges from both societal and scientific perspectives (Gunning and Aha, 2019). From a societal perspective, this opacity undermines trust and accountability, particularly in critical areas such as health, finance, criminal justice, and disaster response, where decisions can significantly impact individuals' lives. It also complicates efforts to ensure fairness and mitigate biases, potentially leading to discriminatory outcomes. In the context of spatial analysis, ensuring spatial fairness and reducing spatial inequalities have recently become a focus as AI is increasingly adopted in applications involving location information (e.g., Xie et al., 2022; Saxena et al., 2024). From a scientific perspective, understanding how a model makes predictions is essential for leveraging AI for scientific discovery beyond just predictive tasks. One of the central questions in spatial analysis has been to understand the spatial processes that generate observed spatial patterns, ultimately answering the question, 'Why are things where they are?' Leveraging AI's capacity to handle large-scale data in a flexible way can enhance the discovery and understanding of spatial processes, but this cannot be achieved without AI explainability (Reichstein et al., 2019). Moreover, the inability to understand model decisions limits debugging and improvement efforts, hindering performance enhancement, which is especially important for the development of geospatial machine learning models. Addressing these issues requires the substantial development of eXplainability AI (XAI) techniques.

The next sections in this chapter will discuss the basics of XAI, its integration and application in spatial analysis. The chapter concludes by highlighting some caveats of XAI and future directions.

## XAI basics and taxonomies

Machine learning has primarily been used for three types of data: text, image, and tabular data. The use of XAI varies significantly depending on the type of data, with each requiring specific techniques to uncover the reasoning behind model predictions. For text data, such as that used in natural language processing (NLP), attention mechanisms are often employed to highlight the words or phrases the model relies on, providing insights into its decision-making process. For image data, methods such as Grad-CAM (Gradient-weighted Class Activation Mapping) and occlusion sensitivity are used to identify key pixels in an image that influence the model's output. These methods are particularly valuable for modeling geospatial data obtained from remote sensing, street view images, and maps (Fu et al., 2024). Hsu and Li (2023) offer a detailed comparison and discussion of image-based XAI methods for geospatial applications. In addition, Cheng et al. (2024) presents a review of XAI's geospatial applications, with a focus on image-related tasks. For tabular data, which is also commonly used in spatial analysis (e.g., census or geo-referenced survey data), feature importance methods that identify important features in the model and partial dependence plots that visualize functional forms of relationships are widely used.

XAI methods can be divided into two main categories: model-based and model-agnostic approaches (Murdoch et al., 2019). Model-based methods are integrated into specific models, offering explanations that are inherently tied to the model's structure. Inherently interpretable models, such as linear regression and decision trees, provide clear explanations through regression coefficients or tree splits. Additionally, model-based methods, such as those used to explain neural networks, are often used in computer vision or NLP tasks. Some feature importance measures for tree-based models, such as those based on impurity or entropy, also fall into this category. On the other hand, model-agnostic methods can be applied to any model, regardless of its structure. The focus of these methods is on the relationship between the input data and the model's output, without relying on the model's internal workings. Examples include permutation feature importance, partial dependence plots, LIME, and Shapley value-based approaches.

In addition to being classified as model-based or model-agnostic, XAI methods can also be distinguished by the level of explanation they provide: global or local (Lundberg et al., 2020). Global XAI methods offer insights into a model's overall behavior, giving a broad view of its decision-making process. These methods are useful for identifying general patterns and rules across a dataset. For example, linear regression coefficients provide a global explanation by showing how each feature affects predictions overall. In more complex models, global explanations often involve examining feature importance or decision rules across the entire dataset such as various global feature importance measures. In recent years, more developments have been focused on local XAI methods which explain the model's behavior for individual predictions. Local XAI offers detailed, case-specific explanations that are particularly important when individuals are of concern rather than groups or the population. More details on local XAI will be introduced in the next section.

## XAI and spatial analysis

The advancement from global to local interpretation provides opportunities for explaining models of geospatial data, where observations are georeferenced by coordinates, addresses, or administrative boundaries, allowing explanations to be visualized as maps (Li, 2022). Among the earliest prominent local XAI methods is Locally Interpretable Model-agnostic Explanations (LIME) by Ribeiro et al. (2016). LIME assumes that any complex model can be approximated as locally linear at a given observation. By creating a simple local surrogate model (e.g., linear regression) from perturbed data drawn from a neighborhood around each observation, it approximates relationships at the local level. However, one of LIME's major challenges is defining local proximity, as results can be highly sensitive to this choice (Molnar, 2020). Additionally, the accuracy of local approximations may vary across observations, complicating the comparison of local coefficients. In an effort to address these issues and make LIME more spatially explicit, Jin et al. (2023) extended the LIME framework by developing the Geographically Localized Interpretable Model-agnostic Explanations (GLIME) method, which defines local neighborhoods based on geographic proximity. They applied GLIME to explain an LSTM model used to classify residential mobility types using both neighborhood- and individual-level characteristics. In this approach, the local explanation model is fixed at the state level, which enables the calculation of state-level feature importance for different feature groups. The results can be visualized on maps to highlight the features that contribute most to predicting residential mobility patterns, which vary across states. GLIME shows promise in capturing state-level effects by sampling data from geographic neighborhoods to account for spatial dependence in each local explanation model. However, it still inherits several limitations from LIME, including the arbitrary selection of neighborhood boundaries, which is subject to the modifiable areal unit problem, and the lack of guarantees for local accuracy and model fidelity. Despite these limitations, GLIME remains a useful tool when interpreting the effects of fixed geographic neighborhoods (e.g., based on administrative boundaries) is of interest.

Another key local XAI development is Shapley value-based approaches from game theory (Shapley, 1953). The Shapley value considers how to fairly distribute contribution among players participating in a coalition game, and the Shapley value for a player $j$ (denoted as $\varphi_j$) in a game is given by:

$$\varphi_j = \sum_{S \subseteq M \setminus \{j\}} \frac{s!\,(p-s-1)!}{p!} \bigl(f(S \cup \{j\}) - f(S)\bigr) \qquad (1)$$

where $p$ is the total number of players, $M\setminus\{j\}$ is a set of all possible combinations of players excluding $j$, $S$ is a player set in $M\setminus\{j\}$ with a size of $s$, $f(S)$ is the outcome of $S$, and $f(S \cup \{j\})$ is the outcome with players in $S$ plus player $i$. The interpretation of Equation (1) is that the Shapley value of a player is the weighted average of its marginal contribution to the game outcome, taken over all possible combinations of players. In the context of machine learning, the players are features, the model is the game, and the prediction represents the outcome. Therefore, Shapley values provide an additive breakdown of individual prediction into marginal contributions from each feature (Štrumbelj and Kononenko, 2014) as shown by Equation 2:

$$\hat{y}_i = \varphi_0 + \sum_{j=1}^{p} \varphi_{ji} \qquad (2)$$

where $\varphi_{ji}$ is the Shapley value for feature $j$ of observation $i$, and $\varphi_0$ is a base value to adjust for the difference between the sum of Shapley values and the prediction value. The Shapley value-based approach has gained significant attention due to the development of Shapley Additive Explanations (SHAP) by Lundberg and Lee (2017) that unifies several XAI methods into one Shapley value framework. Lundberg and Lee (2017) introduced a Shapley-compliant kernel for LIME that enables the two methods to converge on the same solution. The Shapley-compliant kernel is theoretically stronger than LIME, as it maintains local accuracy across all observations. SHAP also incorporates DeepLift for image-related explanations and proposes several estimation methods, such as Tree SHAP, for scalable explanations in tree-based models (Lundberg et al., 2020). SHAP's popularity is enhanced by its user-friendly Python package, *shap* which includes various Shapley value estimation algorithms and visualization techniques. SHAP is widely employed in leading AI platforms, including Google Vertex, IBM Watson OpenScale, Amazon SageMaker, and Microsoft Fabric, and has been applied in numerous domains. Its academic influence is also substantial, with tens of thousands of papers published based on SHAP so far. In spatial studies, SHAP has been widely applied to explain models involving geospatial data to understand how features determine spatial patterns, with applications in areas such as house price modeling (Chen et al., 2020), urban climate analysis (Yuan et al., 2022), species distribution (Song and Estes, 2023), travel behavior (Kashifi et al., 2022; Li, 2023; Zhang et al., 2024), accident analysis (Parsa et al., 2020), disaster prediction (Abdollahi and Pradhan, 2023; Zhang et al., 2023; Cilli et al., 2023; Dikshit et al., 2021), hydrology (Yang et al., 2021; Wang et al., 2022), and public health (Kim and Kim, 2022; Zou et al., 2024), among others.

Although the above-mentioned examples demonstrate the utility of local interpretation methods in geospatial-related applications, it is important to note that most of existing works primarily focus on explaining nonspatial effects, such as global nonlinearity or variable interaction. As a result, the explanations or effects are not explicitly spatial. It worth noting that several studies (e.g., Dou et al., 2023; Wang et al., 2024) have attempted mapping SHAP values to explore the spatial distribution of feature contributions. These maps highlight areas where features have strong/weak and positive/negative contributions to predictions, providing spatial patterns of feature importance. However, it is important to note that SHAP values differ from regression coefficients and should not be interpreted in the same way. Unlike Geographically Weighted Regression (GWR), in which local coefficients directly indicate the sign and strength of the relationship between a feature and the prediction, mapped SHAP values represent the marginal contribution of a feature to a specific prediction and do not imply a directional relationship in the same sense. For example, a positive SHAP value can be observed when a feature's value is below average and there is a negative relationship between feature and prediction. Li (2022) explored further the connection between SHAP and well known spatial statistical models using simulated data and found that when coordinates are input as features, coordinate interactions and their interactions with other features can be analogous to spatial fixed effect and spatially varying effects that have similar coefficient values compared to spatial lag model and multi-scale geographically weighed regression model. However, the off-the-shelf SHAP values have two limitations for spatial data:1) interaction effects can only be calculated for tree-based models not for others such as neural networks; and 2) SHAP does not consider coordinates as a single joint feature which violates Shapley values' properties. To solve both issues, Li (2024) developed GeoShapley that extended Shapley interaction values and Joint Shapley values and proposed a kernel-based estimation method that is agnostic to model structures. More computational details can be found in Li (2024) but the final output of GeoShapley has four components that add up to the model prediction:

$$\widehat{y} = \phi_0 + \phi_{GEO} + \sum_{j=1}^{p} \phi_j + \sum_{j=1}^{p} \phi_{(GEO,j)} \qquad (3)$$

where $\phi_0$ is a constant base value representing the average prediction given the background data, serving as the global intercept. $\phi_{GEO}$ is a vector of size *n* that measures the intrinsic location effect in the model, analogous to the local intercept in geographically weighted regression models or spatial fixed effects. The intrinsic location effect captures unmeasurable contextual influences, which can stem from cultural factors, historical legacies, people's perceptions and attachments to places, and local social interactions, among other factors (Sampson, 2011; Fotheringham and Li, 2023). $\phi_j$ is a vector of size *n* for each non-location feature *j*, representing location-invariant effects, such as global linear and non-linear contributions to the model. $\phi_{(GEO,j)}$ is vector of size *n* for each non-location feature *j*, capturing the spatially varying interaction effects in the model. These effects can be further visualized as spatially varying coefficients, analogous to the coefficients in geographically weighted regression, by estimating the partial derivative between $\phi_{(GEO,j)}$ and $X_j$. If there are no spatial effects in the model, the terms $\phi_{GEO}$ and $\phi_{(GEO,j)}$ will be zero, reducing Equation (3) to Equation (2), which corresponds to the normal SHAP values. GeoShapley thus provides a coherent bridge between traditional statistical approaches and modern machine learning models. Though recent, GeoShapley has been applied in empirical geospatial studies (e.g., Ke et al., 2025; Peng et al., 2025; Foroutan et al., 2025; Wu et al., 2025).

In addition to LIME and Shapley value-based methods, it worth noting there are few other local XAI techniques have been explored or developed within the literature. For instance, Masrur et al. (2022) introduced a spatio-temporally interpretable random forest (iST-RF) that incorporates a spatio-temporal sampling strategy for model training and weighted prediction. This model includes an impurity-based feature importance score to enable both local and global interpretations. The iST-RF outperformed aspatial random forest models in predicting wildfire occurrence. Its sampling strategies also facilitate the construction of local tree models that reveal local feature importance, aiding in the exploration of how environmental variables influence wildfire occurrence and their regional partial relationships. Liu et al. (2024) explored applying GNNExplainer to Graph Neural Network (GNN) models for predicting traffic volume over street networks. GNNExplainer creates counterfactual explanations by perturbing the graph structure to assess changes in predictions. This approach produces local feature importance scores that highlight key node connections and demonstrate how features impact predictions. Given the growing use of GNNs in spatial applications, GNNExplainer offers significant potential for interpreting these complex networks. However, it is worth noting that the feature importance scores in both GNNExplainer and iST-RF are not tied to specific prediction units, making them less intuitive compared to Shapley value-based feature importance scores.

**An example of predicting voting preferences in 2020 Presidential election**

This section demonstrates an empirical example application of XAI in spatial analysis. Given the focus of this book on human geography and GeoAI, voting behavior has been chosen as the example, as studying electoral patterns are key areas of interest in human geography. Here, we model the voting results from the 2020 Presidential election at the county level and examine its determinants using machine learning and GeoShapley. The data is sourced from Fotheringham and Li (2023), who originally obtained the voting data from the MIT Election Lab and socio-demographic data from the 2015-2019 ACS 5-year estimates. The dependent variable is the county-level percentage of votes for the Democratic Party in a two-party contest, excluding third-party votes. Seven key socio-demographic variables (percentage of Black, percentage of Hispanic, percentage of Bachelor's degree of higher, percentage of foreign born, log transformed population density, median household income and percentage of third party vote) were selected to explain the spatial distribution of voting patterns. The analysis covers the contiguous U.S., including 49 states (and Washington, D.C.), for a total of 3,108 counties.

For model training, the Microsoft's FLAML (Fast Library for Automated Machine Learning and Tuning) python package (Wang et al., 2021) was used to train the machine learning models, with hyperparameters optimized through five-fold cross-validation. GeoShapley values were then calculated for the final trained model using the *geoshapley* python package. To estimate the uncertainty of these values, a bootstrap sampling process was applied: 500 bootstrap resamples of the dataset were used to retrain the model, and GeoShapley values were recalculated for each sample. This process generated a distribution of GeoShapley values to account for uncertainties within the machine learning model. To provide a benchmark, a Multi-scale Geographically Weighted Regression (MGWR) model was also estimated using the same data by the *mgwr* python package (Oshan et al., 2019), offering a comparison with a more established spatial model. Code and results reported in this book chapter can be found at this https://github.com/Ziqi-Li/XAI_spatial_analysis_chapter.

In terms of accuracy, the best-performing machine learning model was XGBoost, with a five-fold cross-validated out-of-sample $R^2$ value of 0.933. The MGWR model achieved the same value $R^2$, but it is an in-sample $R^2$. Due to fundamental differences in model estimation and training procedures, these two $R^2$ values cannot be directly compared, but they generally suggest that both models perform similarly well for this dataset. Now, we can first look at the results from XGBoost and GeoShapley. Figure 1 presents a ranking of global feature importance in the XGBoost model based on averaged absolute GeoShapley values across all counties. For each feature, the blue component represents the averaged primary effect, and the red component represents the feature's averaged location-varying effect. The figure shows that the percentage of Black residents and the percentage of Bachelor's degrees have the highest total contribution to the model's prediction, followed by the intrinsic location effect and population density. For each feature, there is a varying amount of location-specific effect, but these effects are smaller than the location-invariant primary effects.

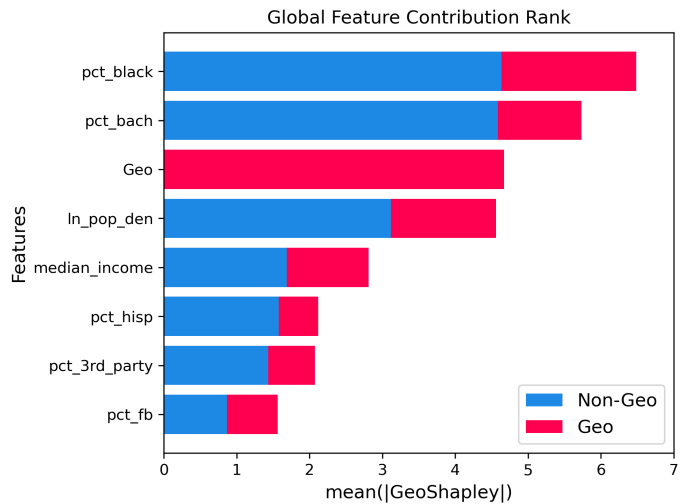

Figure 1. Global feature importance ranking in the XGBoost model based on GeoShapley values.

Figure 2 shows the primary effects using partial dependence plots. The y-axis in each plot represents the contribution of the corresponding feature to the prediction of the Democratic vote share. All features exhibit varying degrees of non-linear association with Democratic vote share. For instance, the percentage of Black population and individuals with a Bachelor's degree or higher show a positive relationship characterized by an S-shaped curve. As the values of these features increase, the Democratic vote share initially increases slowly, then accelerates, before flattening out as the feature values approach their maximum. This indicates non-linear dynamics: at lower percentages of Bachelor's degrees and Black population, the Democratic vote share increases gradually. As feature values increase, the rise in Democratic votes accelerates, reflecting the strong political alignment of more educated and racially diverse populations with the Democratic Party. In counties with high levels of educational attainment and racial diversity, Democratic support grows rapidly, but at very high levels, the increase flattens out again, as the population becomes overwhelmingly Democratic. These non-linear relationships suggest that the effect of education and race on Democratic voting is strongest at moderate levels, while the impact diminishes at very low or very high values. Similar non-linear patterns are observed in other variables: median income has a negative non-linear relationship, while population density (logged), third-party vote share, and the percentage of foreign-born population show a non-linear positive association.

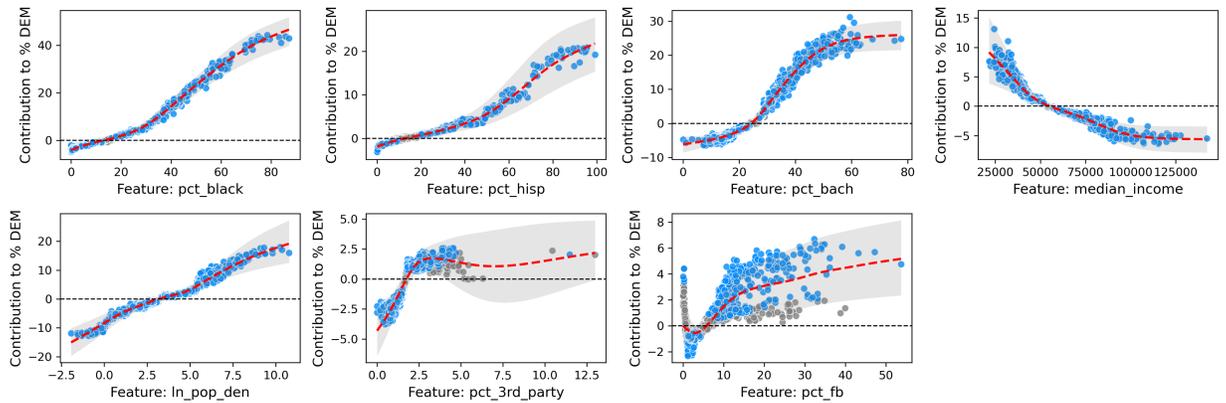

Figure 2. Partial dependence plots of primary effects.

A comparison with the MGWR model is presented below. Li (2024) demonstrated that the contribution from location features is analogous to the local intercept in an MGWR model, both of which measures the intrinsic contextual effects that quantify the importance of location when holding other features constant. Estimates beyond their 95% confidence intervals are masked in grey. The two maps reveal a similar pattern: counties in the South are intrinsically leaning towards the Republican Party, while counties in the Pacific West, Midwest, and Northeast are intrinsically leaning towards the Democratic Party, with other demographic factors held constant. Such pattern aligns well with the general public perception of regional political ideology.

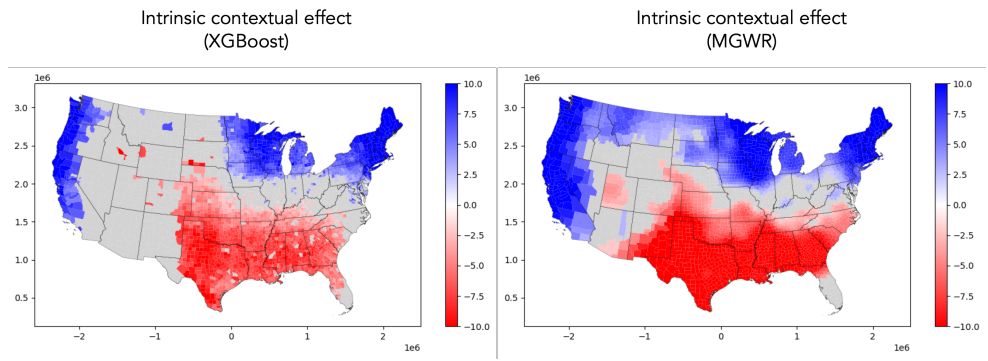

Figure 3. Intrinsic contextual effect estimated from XGBoost + GeoShapley and MGWR.

The comparison between MGWR and XGBoost + GeoShapley spatially varying coefficients for the top three most important features is shown in Figure 4. To obtain spatially varying coefficients from GeoShapley, we can use a spatial smoothing method, such as GWR used here (and is provided in the *geoshapley* package), to regress the combined primary and location-varying effects against the respective feature values. These values are then comparable to MGWR's local coefficients. Figure 4 shows that both approaches reveal similar coefficient patterns. Nationally, for both models, the percentage of Black population, population density, and percentage of Bachelor's degree have a positive relationship with Democratic support. Locally and regionally, for the percentage of Black population, both maps highlight stronger associations with Democratic support in the Mountain West and Southeast, whereas states with smaller Black populations, such as Arizona, New Mexico, and Colorado, show little to no association. For population density, the Northwest Central region shows a stronger positive association compared to the rest of the country. For the percentage of Bachelor's degree feature, the West, Midwest, and Northeast display stronger positive associations compared to other regions, particularly in the Dakotas, where the association is low or insignificant. More detailed interpretations in the voting context can be found in Fotheringham et al. (2021) and Li and Fotheringham (2022).

The comparison demonstrates that XGBoost and MGWR yield similar results in this case study. However, it is worth mentioning that the advantages of adopting a machine learning-based approach are more favorable when: 1) it does not require model specification, allowing the same framework to be applied to datasets generated from processes beyond spatially varying ones; and 2) GeoShapley can separate spatially varying effects and highlight the primary non-linear components, which linear statistical models cannot. On the other hand, statistical approaches are more favorable when the data align with hypotheses and assumptions (e.g., voting is expected to be affected by spatial context), and when formal statistical inference and uncertainty quantification are desired.

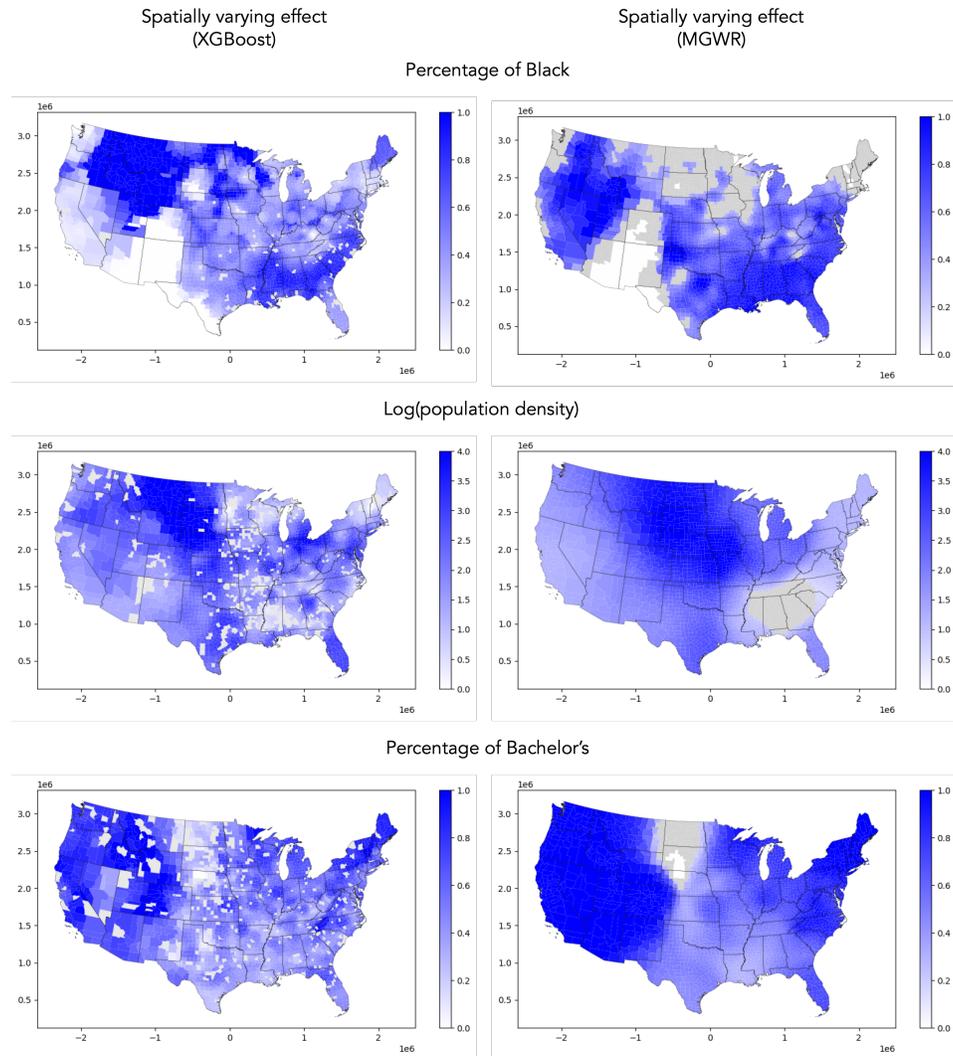

Figure 4. Spatially varying coefficients estimated from XGBoost + GeoShapley and MGWR.

## Challenges and future directions

Despite the success of various XAI methods which provide insights into machine learning models that were previously unavailable, there are several intrinsic limitations, caveats or challenges. The most important aspect is that XAI describes model behavior but not the true data-generating processes. This has several implications. First, XAI should not be used if a model performs poorly, which may be due to poor-quality data, such as small data size, feature uncertainties, missing important features, insufficient feature engineering, biased data, missing data, among other issues. Poor model performance may also result from overfitting, underfitting, inadequate tuning, or model-data incompatibility. In any of these cases, XAI explanations may not be sensible or useful. Substantial model training and feature engineering efforts are required to develop a robust model before generating good explanations. Furthermore, despite the success of GeoAI in image-related tasks, GeoAI models for traditional tabular data regression and classification problems are still in their infancy, which is arguably one of the most important tasks in spatial analysis. More work is needed to develop robust and transferable GeoAI models for tabular data tasks accounting for unique characteristics of spatial tabular data. Finally, computing uncertainties for XAI explanations will help quantify the uncertainties in the model. For the example presented above, a bootstrap strategy can be adopted to generate a distribution of GeoShapley values, from which confidence intervals can be derived. The second implication is that model explanations can vary between different models, especially when features are correlated. For example, Shpapley values for Random Forest and XGBoost can differ because of how each model handles feature correlations (Chen et al., 2020; Li, 2022). In Random Forest, feature importance is often spread more evenly across correlated features (Strobl et al., 2007), while in XGBoost, one feature may dominate the explanation due to its sequential learning process (Chen and Guestrin, 2016). Although Shpapley values still reflect what a model considers important, special attention is needed when interpreting and reporting the results.

One caveat worth stressing is in interpreting the explanations derived from Shapley values. Shapley values represent the marginal contribution of features to the model prediction while holding all other features constant. Though there is a direct link between Shapley values and linear regression coefficients, Shapley values themselves do not have the same interpretation as regression coefficients. Positive Shapley values can result from lower-than-average feature values while having a negative relationship with prediction or higher-than-average feature values while having a positive relationship with prediction. This can cause confusion, especially when visualizing Shapley values spatially, as it involves visualizing four-dimensional spatial partial dependence, which is inherently more challenging than aspatial effects like those shown in Figure 2. GeoShapley attempts to use GWR as a spatial smoother to generate a spatially varying coefficients, but these are inevitably exploratory and deviate from the true GeoShapley values which adhere to game-theory properties. Future work should focus on better and more formal methods for visualizing spatial contributions in models.

Additionally, although a significant body of literature in GeoAI has focused on improving model performance, equally important is the development of more spatially explicit interpretation methods to better understand geographic knowledge from machine learning models. GLIME by Jin et al. (2023) and GeoShapley by Li (2024) are good examples, and they inspire further developments to highlight spatial effects that spatial data scientists are interested in.

Lastly, there has been increasing interest in spatial causality to better understand the 'why' question, going beyond correlation analysis (Akbari, Winter, and Tomko, 2023; Credit and Lehnert, 2023; Gao et al., 2023; Zhang and Wolf, 2023). Janzing, Minorics, and Blöbaum (2020) argued that Shapley values inherently have a causal interpretation when using interventional algorithms (such as Kernel SHAP and GeoShapley), which examine how a model behaves when the dependence among features is deliberately disrupted. Shapley values also provide a way to investigate more complex causal effects within machine learning models and have been shown to be effective in measuring heterogeneous treatment effects using instrumental variables (Syrgkanis et al., 2019). Furthermore, Heskes et al. (2020) introduced causal Shapley values, a model-agnostic approach to measure the direct and indirect effects of each feature on the model's prediction. Extending and applying these causal ML methods to geospatial tasks would be a valuable direction for future research.

# References

Anselin, L. (1988). *Spatial econometrics: methods and models*(Vol. 4). Springer Science & Business Media.

Biljecki, F. (2023). GeoAI for Urban Sensing. In *Handbook of Geospatial Artificial Intelligence* (pp. 351-366). CRC Press.

Brunsdon, C., Fotheringham, A. S., & Charlton, M. E. (1996). Geographically weighted regression: a method for exploring spatial nonstationarity. *Geographical analysis*, *28*(4), 281-298

Chen, H., Janizek, J. D., Lundberg, S., & Lee, S. I. (2020). True to the model or true to the data?. *arXiv preprint arXiv:2006.16234*.

Chen, T., & Guestrin, C. (2016, August). Xgboost: A scalable tree boosting system. In *Proceedings of the 22nd acm sigkdd international conference on knowledge discovery and data mining* (pp. 785-794).

Cheng, X., Vischer, M., Schellin, Z., Arras, L., Kuglitsch, M. M., Samek, W., & Ma, J. (2024). Explainability in GeoAI. In *Handbook of Geospatial Artificial Intelligence* (pp. 177-200). CRC Press.

Dou, M., Gu, Y., & Fan, H. (2023). Incorporating neighborhoods with explainable artificial intelligence for modeling fine-scale housing prices. *Applied Geography*, *158*, 103032.

Foroutan, E., Hu, T., & Li, Z. (2025). Revealing key factors of heat-related illnesses using geospatial explainable AI model: A case study in Texas, USA. *Sustainable Cities and Society*, 122, 106243.

Fotheringham, A. S., & Li, Z. (2023). Measuring the unmeasurable: models of geographical context. *Annals of the American Association of Geographers*, *113*(10), 2269-2286.

Fotheringham, A. S., Brunsdon, C., & Charlton, M. E. (2002). Geographically weighted regression. *The Sage handbook of spatial analysis*, *1*, 243-254.

Fotheringham, A. S., Oshan, T. M., & Li, Z. (2023). *Multiscale geographically weighted regression: Theory and practice*. CRC Press.

Fu, C., Zhou, Z., Xin, Y., & Weibel, R. (2024). Reasoning cartographic knowledge in deep learning-based map generalization with explainable AI. *International Journal of Geographical Information Science*, 1-22

Gao, S., Hu, Y., & Li, W. (Eds.). (2023). *Handbook of geospatial artificial intelligence*. CRC Press.

Griffith, D. A. (2003). *Spatial filtering* (pp. 91-130). Springer Berlin Heidelberg.

Gunning, D., & Aha, D. (2019). DARPA's explainable artificial intelligence (XAI) program. *AI magazine*, *40*(2), 44-58.

Helbing, D. (2010). *Quantitative sociodynamics: stochastic methods and models of social interaction processes*. Springer Science & Business Media.

Hengl, T., Heuvelink, G. B., & Rossiter, D. G. (2007). About regression-kriging: From equations to case studies. *Computers & geosciences*, *33*(10), 1301-1315.

Hsu, C. Y., & Li, W. (2023). Explainable GeoAI: can saliency maps help interpret artificial intelligence's learning process? An empirical study on natural feature detection. *International Journal of Geographical Information Science*, *37*(5), 963-987.

Jin, C., Park, S., Ha, H. J., Lee, J., Kim, J., Hutchenreuther, J., & Nara, A. (2023). Predicting households' residential mobility trajectories with geographically localized interpretable model-agnostic explanation (GLIME). *International Journal of Geographical Information Science*, *37*(12), 2597-2619.

Lee, D. (2013). CARBayes: an R package for Bayesian spatial modeling with conditional autoregressive priors. *Journal of Statistical Software*, *55*(13), 1-24.

Lek, S., Delacoste, M., Baran, P., Dimopoulos, I., Lauga, J., & Aulagnier, S. (1996). Application of neural networks to modelling nonlinear relationships in ecology. *Ecological modelling*, *90*(1), 39-52.

Li, W., Hsu, C. Y., & Hu, M. (2021). Tobler's First Law in GeoAI: A spatially explicit deep learning model for terrain feature detection under weak supervision. *Annals of the American Association of Geographers*, *111*(7), 1887-1905.

Li, W. (2020). GeoAI: Where machine learning and big data converge in GIScience. *Journal of Spatial Information Science*, (20), 71-77.

Li, Z. (2022). Extracting spatial effects from machine learning model using local interpretation method: An example of SHAP and XGBoost. *Computers, Environment and Urban Systems*, *96*, 101845.

Li, Z. (2023). Leveraging explainable artificial intelligence and big trip data to understand factors influencing willingness to ridesharing. *Travel Behaviour and Society*, *31*, 284-294.

Li, Z. (2024). GeoShapley: A Game Theory Approach to Measuring Spatial Effects in Machine Learning Models. *Annals of the American Association of Geographers*, 1-21.

Li, Z., & Fotheringham, A. S. (2022). The spatial and temporal dynamics of voter preference determinants in four US presidential elections (2008–2020). *Transactions in GIS*, *26*(3), 1609-1628.

Lindgren, F., & Rue, H. (2015). Bayesian spatial modelling with R-INLA. *Journal of statistical software*, *63*(19).

Liu, P., Zhang, Y., & Biljecki, F. (2024). Explainable spatially explicit geospatial artificial intelligence in urban analytics. *Environment and Planning B: Urban Analytics and City Science*, *51*(5), 1104-1123.

Lundberg, S. (2017). A unified approach to interpreting model predictions. *arXiv preprint arXiv:1705.07874*.

Lundberg, S. M., Erion, G., Chen, H., DeGrave, A., Prutkin, J. M., Nair, B., ... & Lee, S. I. (2020). From local explanations to global understanding with explainable AI for trees. *Nature machine intelligence*, *2*(1), 56-67.

Murdoch, W. J., Singh, C., Kumbier, K., Abbasi-Asl, R., & Yu, B. (2019). Definitions, methods, and applications in interpretable machine learning. *Proceedings of the National Academy of Sciences*, *116*(44), 22071-22080.

Oshan, T. M., Li, Z., Kang, W., Wolf, L. J., & Fotheringham, A. S. (2019). mgwr: A Python implementation of multiscale geographically weighted regression for investigating process spatial heterogeneity and scale. *ISPRS International Journal of Geo-Information*, *8*(6), 269.

Peng, Z., Ji, H., Yuan, R., Wang, Y., Easa, S. M., Wang, C., ... & Zhao, X. (2025). Modeling and spatial analysis of heavy-duty truck CO2 using travel activities. *Journal of Transport Geography*, 124, 104158.

Ke, E., Zhao, J., & Zhao, Y. (2025). Investigating the influence of nonlinear spatial heterogeneity in urban flooding factors using geographic explainable artificial intelligence. *Journal of Hydrology*, 648, 132398.

Rial, J. A., Pielke, R. A., Beniston, M., Claussen, M., Canadell, J., Cox, P., ... & Salas, J. D. (2004). Nonlinearities, feedbacks and critical thresholds within the Earth's climate system. *Climatic change*, *65*, 11-38.

Ribeiro, M. T., Singh, S., & Guestrin, C. (2016, August). " Why should i trust you?" Explaining the predictions of any classifier. In *Proceedings of the 22nd ACM SIGKDD international conference on knowledge discovery and data mining* (pp. 1135-1144).

Sachdeva, M., Fotheringham, A. S., Li, Z., & Yu, H. (2022). Are We Modelling Spatially Varying Processes or Non‐linear Relationships?. *Geographical Analysis*, *54*(4), 715-738.

Sampson, R. J. (2011). 11 Neighborhood effects, causal mechanisms and the social structure of the city. *Analytical sociology and social mechanisms*, 227.

Saxena, N. A., Zhang, W., & Shahabi, C. (2024). Spatial Fairness: The Case for its Importance, Limitations of Existing Work, and Guidelines for Future Research. *arXiv preprint arXiv:2403.14040*.

Song, L., & Estes, L. (2023). ITSDM: Isolation forest‐based presence‐only species distribution modelling and explanation in R. *Methods in Ecology and Evolution*, *14*(3), 831-840.

Sonoda, S., & Murata, N. (2017). Neural network with unbounded activation functions is universal approximator. *Applied and Computational Harmonic Analysis*, *43*(2), 233-268.

Stewart Fotheringham, A., Li, Z., & Wolf, L. J. (2021). Scale, context, and heterogeneity: A spatial analytical perspective on the 2016 US presidential election. *Annals of the American Association of Geographers*, *111*(6), 1602-1621.

Strobl, C., Boulesteix, A. L., Zeileis, A., & Hothorn, T. (2007). Bias in random forest variable importance measures: Illustrations, sources and a solution. *BMC bioinformatics*, *8*, 1-21.

Wang, C., Wu, Q., Weimer, M., & Zhu, E. (2021). Flaml: A fast and lightweight automl library. *Proceedings of Machine Learning and Systems*, *3*, 434-447.

Wang, N., Zhang, H., Dahal, A., Cheng, W., Zhao, M., & Lombardo, L. (2024). On the use of explainable AI for susceptibility modeling: Examining the spatial pattern of SHAP values. *Geoscience Frontiers*, *15*(4), 101800.

Wang, S., Huang, X., Liu, P., Zhang, M., Biljecki, F., Hu, T., ... & Bao, S. (2024). Mapping the landscape and roadmap of geospatial artificial intelligence (GeoAI) in quantitative human geography: An extensive systematic review. *International Journal of Applied Earth Observation and Geoinformation*, *128*, 103734.

Wang, S., Peng, H., & Liang, S. (2022). Prediction of estuarine water quality using interpretable machine learning approach. *Journal of Hydrology*, *605*, 127320.

Wu, R., Yu, G., & Cao, Y. (2025). The impact of industrial structural transformation in the Yangtze River economic belt on the trade-offs and synergies between urbanization and carbon balance. *Ecological Indicators*, 171, 113165.

Xie, Y., He, E., Jia, X., Chen, W., Skakun, S., Bao, H., ... & Ravirathinam, P. (2022, June). Fairness by "where": A statistically-robust and model-agnostic bi-level learning framework. In *Proceedings of the AAAI Conference on Artificial Intelligence* (Vol. 36, No. 11, pp. 12208-12216).

Xing, J., & Sieber, R. (2023). The challenges of integrating explainable artificial intelligence into GeoAI. *Transactions in GIS*, *27*(3), 626-645.

Yang, Y., & Chui, T. F. M. (2021). Modeling and interpreting hydrological responses of sustainable urban drainage systems with explainable machine learning methods. *Hydrology and Earth System Sciences*, *25*(11), 5839-5858.

Zhang, X., Zhou, Z., Xu, Y., & Zhao, X. (2024). Analyzing spatial heterogeneity of ridesourcing usage determinants using explainable machine learning. *Journal of Transport Geography*, *114*, 103782.